# Temperature Estimation in Induction Motors using Machine Learning

Dinan Li, Panagiotis Kakosimos
*ABB AB Corporate Research*
Västeras, Sweden
Email: {dinan.li, panagiotis.kakosimos}@se.abb.com

*Abstract*—The number of electrified powertrains is ever increasing today towards a more sustainable future; thus, it is essential that unwanted failures are prevented, and a reliable operation is secured. Monitoring the internal temperatures of motors and keeping them under their thresholds is an important first step. Conventional modeling methods require expert knowledge and complicated mathematical approaches. With all the data a modern electric drive collects nowadays during the system operation, it is feasible to apply data-driven approaches for estimating thermal behaviors. In this paper, multiple machine-learning methods are investigated on their capability to approximate the temperatures of the stator winding and bearing in induction motors. The explored algorithms vary from linear to neural networks. For this reason, experimental lab data have been captured from a powertrain under predetermined operating conditions. For each approach, a hyperparameter search is then performed to find the optimal configuration. All the models are evaluated by various metrics, and it has been found that neural networks perform satisfactorily even under transient conditions.

*Index Terms*— Induction motor, machine learning, temperature monitoring, neural network.

## I. Introduction

There are various reasons that cause a powertrain to fail. In [1], the percentage share of specific failures in induction machines has been presented. According to this study, more than 40% of failures are due to bearing malfunctions, while 38% are issues in the machine stator. Several of the listed failures are attributed to excessive heat. The windings of the stator and rotor as well as the bearings are vital components that are susceptible to high temperatures. Therefore, water- or air-cooling techniques have been developed to cool the powertrain components. Plenty of techniques have been created for monitoring the temperatures in the interior of motors and drives. The present methods in the literature can be classified into three main categories by the used method: (1) sensor-based, (2) model-based, and (3) artificial-intelligence (AI) algorithm-based [2]. All the various methods exhibit different advantages and disadvantages; however, AI algorithm-based methods are becoming more attractive nowadays.

Machine-learning (ML) has enhanced the application of more complicated data-driven modeling approaches on edge devices without powerful processors or cloud computing services with minimal data transfer needs. A number of recent works in the literature use ML for monitoring the performance of electric powertrains, including storage means, machines, and power converters [3]–[5]. More specifically, deep NNs have been used in [6], [7], to predict the temperatures of the permanent magnets in an automotive application. The magnet temperatures were monitored accurately by the offline-trained model under different operating scenarios. In [8], a method of estimating the dynamic thermal map of a power chip and the rate of temperature change with neural networks (NNs) was proposed. The junction temperature of an IGBT power module was estimated in [9] by employing three non-linear ML-based models that offered a great prospect for industrial applications.

Furthermore, in [10], ML was combined with traditional thermal modeling approaches, such as Lumped Parameter Thermal Networks (LPTNs). The adaptation of the network parameters enabled the model to enhance its performance and take environmental and aging phenomena into account. In [11], the battery health was supervised by considering numerous different charging profiles during the training of the ML model. Temperature differences enhanced the performance of NNs in [12], whereas in [13], a multi-layer perceptron (MLP) model was developed to estimate the winding temperatures of a brushless DC motor. It is apparent that many benefits originate from the use AI/ML approaches; thus, their applicability is being extended. The knowledge of temperatures at different locations in a powertrain is the enabler of different use cases such as condition monitoring and control optimization.

This paper concentrates on developing an ML model for estimating internal motor temperatures. Therefore, several models have been designed and trained. For enabling the model deployment to commercial applications, only signals available in modern electric drives have been considered. The emphasis has been on those solutions that take continuous target variables as inputs due to the nature of the temperature signals. Linear regression, multilayer perception (MLP), and convolutional neural networks (CNNs) have been analyzed, and their performances in estimating internal temperatures of electric motors have been evaluated. An experimental test bench has been used to implement and train the examined models for temperature prediction under varying operating conditions. The developed ML models have been able to estimate internal temperatures with high precision, opening up the prospects of implementing similar models to commercial applications.

## II. Machine Learning Approaches

Machine learning embodies a wide range of methods that help create models that emulate the behaviors of systems accurately. Those methods identify patterns from historical information, then apply the discovered patterns to estimate future behaviors. ML algorithms can be classified into four classes: (1) supervised, (2) unsupervised, (3) semi-supervised, and (4) reinforcement learning [14]. Supervised learning algorithms are the focus of this work as data from the historical performance are available. They can also be divided further into two groups: (1) classification or (2) regression. Classification deals with classes or discrete values as outputs, whereas regression needs continuous targets. Temperature signals are continuous targets, thus requiring regression approaches to be used.

*A. Loss function*

A loss or cost function is a function that is used to characterize how well a model fits the data. The loss function is calculated by considering estimations and measurements [15]. Many candidate loss functions exist in the literature classified by the nature of the estimated targets. For instance, the loss functions in classification differ from those in regression problems. In the latter, the mean squared error (MSE) is the most prevalent function. Its mathematical formulation is as next:

$$\text{MSE} = \frac{1}{N}\sum_{i=1}^{N}(Y_i - \widehat{Y}_i)^2. \quad (1)$$

where $N$ is the number of samples, $Y$ is the measured value, and $\widehat{Y}$ is the predicted one by the model. The loss functions are mainly used in the training process, where all the tunable model parameters are optimized by minimizing the total error. Gradient descent has been used in this work because of its effectiveness [16]. A gradient points to the direction where the function increases the most; thus, the tunable parameters are updated in the direction opposite to the gradient decreasing the loss at the fastest pace. The main formula is as below:

$$a_{n+1} = a_n - \gamma \nabla L(a_n), \quad (2)$$

where $a$ is the tunable parameter to be updated, $L$ is the loss function to be minimized, and $\gamma$ impacts the step size of each update per each step. The batch, stochastic, and mini-batch gradient descents are the three main variants. One of their differences is found in the amount of data each algorithm requires for each update. These variants accelerate the optimization process, but they also affect the algorithm convergence. Selecting a proper schedule for the adaptation of the learning rate is thus essential.

*B. Linear regression*

Linear regression is the simplest form of machine learning [17]. This approach considers the existence of a linear relationship between the scalar response and explanatory variables. More specifically, if we consider $n$ observations $\{X_i, y_i\}_{i=1}^{n}$ where $X_i$ is a $n$ by $p$ matrix, the linear relationship can be described as follows:

$$y_i = \beta_0 + \beta_1 x_{i1} + \beta_2 x_{i2} + \cdots + \beta_p x_{ip} + \epsilon_i, \quad (3)$$

where the term $\epsilon$ is used to model the unobserved random errors. The target of this approach is to determine the weights and bias $\beta$ that minimize the loss function. The mean squared error may be used for comparing the measured and predicted values. Due to the simplified nature of the method, the minimization of the mean squared error ends up to a closed-form mathematical solution without a gradient descent. This is commonly referred as the least square method, where $\beta$ is found as follows:

$$\widehat{\beta} = (X^{TX})^{-1}X^T Y. \quad (4)$$

The least square method gives an exact solution to the problem; however, it is not usually preferred because it is computationally intensive when having the large data sizes. The gradient descent method is applied as an alternative. In many applications, the trained model is preferred to be simple meaning that the model is not required to learn so many details from the data. This prevents overfitting and improves the generalizability of the model [18]. In linear regression, regularization helps achieve this goal by adding penalty terms to the loss function and restricting the weights $\beta_i$ from being too large. In particular, an elastic net regularization corresponds to the following:

$$\beta = argmin_\beta \left( ||y - X\beta||^2 + \lambda_2 ||\beta||^2 + \lambda_1 ||\beta||_1 \right). \quad (5)$$

where the first term is the original mean squared error, the second quadratic penalty term is used in ridge regression, and the final term in lasso regression. Although lasso regression is primarily used for sparsification, elastic net can combine it with ridge regression thus taking advantage of both methods.

The loss function can take alternative forms for reaching different results. If the function remains differentiable, gradient descent can be used in the optimization. Other common loss functions that are extensively applied are the *epsilon* insensitive loss function and the *Huber* loss function. The *epsilon* insensitive loss function does not penalize errors within a margin of tolerance. Beyond this margin, the common mean absolute error is used to determine the errors. This is referred to as hinge loss, and when it is combined with linear regression, the resulting model is the support vector regression:

$$L = \begin{cases} 0, & |y - f(X, \beta)| \leq \epsilon \\ |y - f(X, \beta)| - \epsilon, & otherwise. \end{cases} \quad (6)$$

This equation shows the *epsilon* insensitive loss where $\epsilon$ is a tunable parameter. However, the real strength of a support vector regression comes with the kernel method which can transform the approach to handle nonlinear behaviors. The algorithm projects the data into higher dimensional spaces, and then uses hyperplanes to approximate them.

The *Huber* loss is an alternative loss function that is used in robust regression and is less sensitive to outliers in the data set. It performs better than the mean squared loss function when the data set contains a lot of outliers. It has the next form:

$$L = \begin{cases} \frac{1}{2}(y - f(X, \beta))^2, & |y - f(X, \beta)| \leq \delta \\ \delta\left(y - f(X, \beta) - \frac{1}{2}\delta\right), & otherwise. \end{cases} \quad (7)$$

*C. Multilayer perception and backpropagation*

A multilayer perception (MLP), or fully connected neural network, is the simplest kind of feed-forward neural network. This type of network usually includes three layers at least: an input layer, a hidden layer, and an output layer (see Fig. 1). Each layer contains several neurons. All the neurons between two successive layers are connected in a fully connected neural network. Neurons that are found in next layers are linear combinations of the previously connected neurons after applied an activation function:

$$f(X; W_j) \coloneqq \sigma\left(w_{0j} + \sum_{i=1}^{N} w_{ij} x_i\right), \quad (8)$$

where $w_{0j}$ is a bias term for each neuron, $\sigma(\cdot)$ is the activation function that can have various forms. Common functions are sigmoid, tanh, and ReLU [19]. The activation functions are important because they change the model nature to nonlinear. Despite the fact that MLP with only a single hidden layer can estimate any given targets, stacking more layers improves the network generalizability.

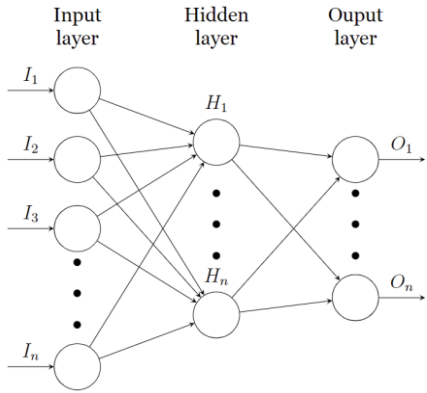

Fig. 1. A multilayer perception (fully connected neural network).

The process of training a neural network frequently consists of two passes. In the first pass, the information is propagated from the input to the output, and the computations are executed (forward pass). The results calculated by the neural network and the actual targets are then compared by using the adopted loss function. Afterward, the gradients are computed from the output layer to the input layer with the chain rule (backpropagation). All the weights and biases of the network are updated.

### D. Convolutional neuron network

The targets in this work are sequential in the form of temperatures; thus, neural networks that receive sequential data can potentially perform better than simplified fully connected neural ones. Convolutional neuron networks are thus good candidates. While 2D convolutional neuron networks are used in image processing applications due to their capability to explore spatial data, the 1D variant naturally is a good candidate for exploring temporal data [20]. Fig. 2 shows the basic structure of a such network with two layers. The first layer has a filter size of three meaning that three neurons from the previous layer are connected to the filter to produce one output. This filter is reused in the whole layer. The second layer has a filter size of two for producing the output. The filter linearly combines all the inputs with the corresponding weights. For making the network nonlinear, an activation function is assigned to each neuron.

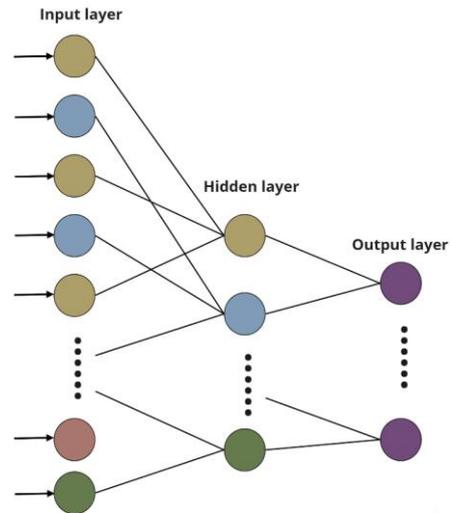

Fig. 2. Basic structure of a one-dimensional convolutional neural network.

Although the structures in Fig. 1 and Fig. 2 seem to be similar, their input layers stand for different meanings. For an MLP, a corresponding output is generated in one time step. Each neuron in the input layer represents one dimension (feature) of the input data. For a temporal CNN, however, the data comes in sequential form meaning that one neuron at the input layer represents the whole input data at one timestep regardless of its dimension. Adjacent neurons represent inputs from adjacent timesteps. Each neuron not only has information from the current time step, but it also has information from previous inputs. If more convolutional layers are stacked, neurons at the latter layers get information from inputs that originate from a long time in the past.

### III. ML MODEL DEVELOPMENT

In this work, a linear model, a fully connected multi-layer perceptron, and a temporal convolutional neural network are investigated further.

### A. Linear model

A linear model can be viewed as the baseline in this paper. Although it is a simple model, it has the potential to be applied to real-world applications (Fig. 3). A class called SGD Regressor from the *scikit-learn* library has been employed to develop different linear models. The search space includes the regularization multiplier parameter, mixing parameter as well as loss functions. With some loss functions, the SGD Regressor can transform the linear model into a nonlinear one, for example, by specifying a kernel with *epsilon* insensitive loss function. However, this possibility is not explored in the hyperparameter search space.

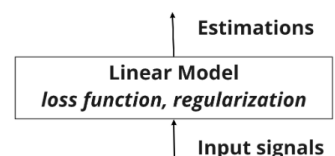

Fig. 3. Block diagram of a simple linear model.

## B. Multi-Layer Perception

MLP is the simplest neural network structure (Fig. 4); however, the used activation function makes it more complicated than linear regression and increases its size. The MLP takes the same input shape as linear models. The hyperparameter search space now becomes the number of neurons on each layer, the number of layers, and the dropout ratio.

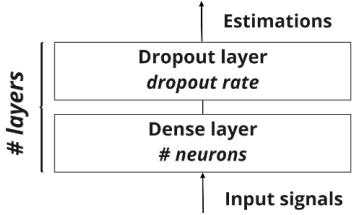

Fig. 4. Block diagram of a fully connected network.

## C. Convolutional Neural Networks

In contrast to the previous two models, input data must be prepared in a sequential form for the CNN model to work (Fig. 5). The model predicts the output at a one-time step in the future by takes a sequence of input data. The sequence length is a hyperparameter that needs to be fine-tuned. To reduce a sequence of data needed for one prediction, the *Global Pooling* layer downsamples the inputs horizontally the dense layer performs the regression.

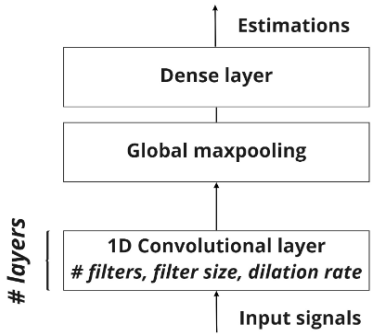

Fig. 5. Block diagram of a convolutional neural network.

In the image processing context, the CNN layers are often used as feature extraction layers, whereas the dense layer performs the regression or classification based on the extracted features. CNN layers can also be stacked, so the number of layers and filters, filter sizes on each layer, the dropout and dilation rates can all be defined based on the application needs.

## D. Evaluation metrics

After defining the hyperparameters of each model, the training process takes place. In the process of training and testing, there are three different losses: training loss, validation loss, and testing loss. The training loss is calculated on the training set for each batch of data and is continuously updated during the training process. The validation loss is calculated on a validation set for each epoch, whereas the set of data is separated from the training set before the training process starts. This loss is employed to assess the model performance during the training time and is used as a checkpoint for storing the optimal models and deciding the early stopping. After finishing the training, the model is then evaluated on a final testing set that is also separated from the training and validation sets. The testing loss is employed to assess the final model performance.

Various metrics are used to assess the performance. Mean squared error (MSE) is the most commonly used, as already mentioned. The Mean Absolute Error (MAE) is very similar to MSE with the mere difference of replacing the square with the absolute value. Mean absolute error gives a more interpretable result than MSE. Although the error function is not differentiable in the whole space due to the absolute value, it can be used as a loss function with sub-gradient descent methods. On the other hand, the MSE loss function has a particular advantage over the MAE one because it penalizes large errors heavier. Since large deviations in temperatures must be avoided, this loss function is a good candidate.

L-infinity finds the largest error for a set of estimations and is significant in temperature estimation for keeping the largest error small. Some ML models may not have low MSE or MAE, but if they have good L-infinity numbers, they remain good candidates. R-squared, also known as the coefficient of determination, is a metric often used in evaluating the performance of regression models. In a general, there is a data set $\{y_1, y_2, y_3, ..., y_i\}$ with a mean of $\bar{y}$ and an estimation dataset as $\{\widehat{y_1}, \widehat{y_2}, \widehat{y_3}, ..., \widehat{y_i}\}$. Then, two sums of squares formulas can be defined. The sum of squares of residuals:

$$SS_{\text{res}} = \sum_i (y_i - \hat{y}_i)^2 = \sum_i e_i^2 \qquad (9)$$

and the total sum of squares:

$$SS_{\text{tot}} = \sum_i (y_i - \bar{y})^2. \qquad (10)$$

Then, R-squared is defined as follows:

$$R^2 = 1 - \frac{SS_{\text{res}}}{SS_{\text{tot}}}. \qquad (11)$$

The coefficient of determination measures the percentage of variance that can be explained by the regression model with respect to the total variance. It provides a normalized metric between zero and one.

## IV. SYSTEM SPECIFICATIONS

This chapter gives a detailed step-by-step implementation of the work that has been done. The methods described in the previous chapter will be applied during implementation.

### A. Collection of the training datasets

A lab setup has been used to generate the training datasets and evaluate the performances of the developed ML models. Table I summarizes the specifications of the used induction motor in the experimental test bed. Different operating profiles have been generated and applied to the motor. A profile defines the operating conditions and includes both the speed and torque of the motor. To create a variety of profiles, the dynamics and rate of change have been varied and separated into three groups: slow, medium, and fast. In total, eighteen profiles were generated giving approximately 150 hours of raw data.

TABLE I. MOTOR SPECIFICATIONS

| Motor | Test motor |
|---|---|
| Type | Induction machine |
| Power rating | 15 kW |
| Voltage rating | 400 V |
| Current rating | 30.6 A |
| Torque rating | 97 N m |
| Pole pair number | 2 |
| Speed | 1478 rpm |
| Cooling | Forced air |

Temperature sensors mounted on and inside the motor have been used to measure the temperature of the windings and bearings. Several modern motors are already equipped with such sensors. The temperature data have been collected with a sampling frequency of 1 Hz. In parallel, data already collected by the drive are also stored at a rate of 1 Hz. This includes all the parameters of interest during the operation of a motor such as current, voltage, speed, and more.

B. *Data selection and preprocessing*

During the motor operation, several signals have been captured along with the targeted temperatures of windings and bearings. The correlation matrix showed that the motor speed and current were related the most to the thermal behavior of the motor. Additionally, the motor shell temperature, or reference temperature, is important because it provides the boundary conditions of the thermal model and the context of the environmental conditions. Table II summarizes all the inputs and outputs of the model.

TABLE II. MODEL INPUTS AND OUTPUTS

| Parameter | Symbol |
|---|---|
| Measured inputs | |
| Motor speed | $n_m$ |
| Motor current | $I_m$ |
| Reference temperature | $T_{ref}$ |
| Target temperatures | |
| Winding | $T_W$ |
| DE bearing | $T_{DE}$ |
| NDE bearing | $T_{NDE}$ |

The next step is to normalize the model inputs by transforming each feature into a standard normal distribution. Data following such distribution have a mean equal to zero and a unit variance. This step is especially important because it makes the input of a network contain both positive and negative values, which in turn accelerates the learning process. In addition, it also transforms distinct features into similar ranges, and the network can learn from each feature with a similar effort.

The last pre-processing task is the feature expansion with exponentially weighted moving averages for each selected feature at each time step. In this way, linear models, or models like MLP that are not originally designed for handling sequential data have sufficient references to past motor behavior. On the other hand, the performance of CNNs is also improved by minimizing the window length. Since the models need to have both short- and long-term memories, eight span values ranging have been considered.

C. *Hyperparameter search*

With all the necessary information available, the models are ready to be trained. Before finalizing the parameters of each model, it is essential to run a hyperparameter search. During this process, a lot of parameters are fine-tuned. The finally selected model configurations have been summarized in Table III.

TABLE III. MODEL CONFIGURATION

| Symbol | Hyperparameter | Value |
|---|---|---|
| | Linear model | |
| | Penalty coefficient | 0.43 |
| $l1$ | Mixing parameter | 0.99 |
| $loss$ | Loss function | Squared error |
| | MLP (2 layers) | |
| $n$ | No. neurons | 90,20 |
| | CNN (3 layers) | |
| $n_{filter}$ | No. filters | 125,5,125 |
| $s_{filter}$ | Filter sizes | 2,2,2 |
| $d$ | Dilation rates | 3,1,1 |
| $l$ | Sequence length | 100 |

To test the model performance thoroughly, all the generated profiles have been used as test data at least once. This has been achieved by removing each profile from the training set one by one and then training the model on the rest of the profiles. During training, another small set of profiles has been taken away as the validation set.

V. RESULTS AND DISCUSSION

In this section, the three investigated ML models with their best configurations from the hyperparameter search have been used to predict the motor temperatures under one representative operating profile. As shown in Fig. 6, the selected profile includes both slow and fast dynamics making it a good example for assessing the model performances.

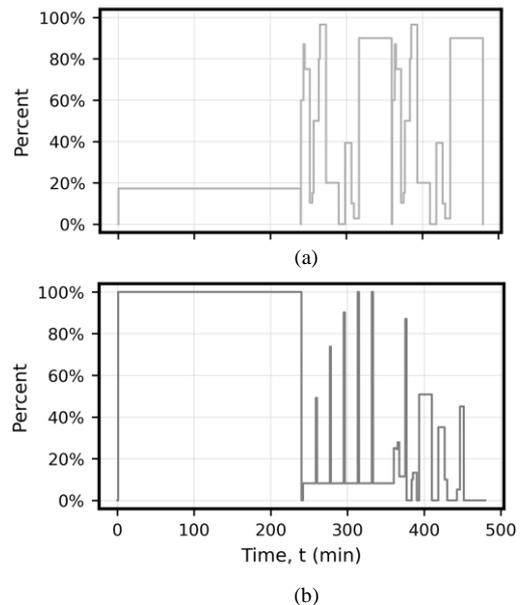

Fig. 6. Motor torque (a) and speed (b) under the selected operating profile.

A. *Linear model*

The hyperparameter search selected a penalty coefficient of 0.43 and a mixing parameter of 0.99. The model is L1 penalized because of the high mixing parameter. Fig. 7 gives a summary of the performance of the linear model trained to estimate the motor targets. The first row of the figure shows the measured actual temperature and the predicted temperature. The second row shows the corresponding errors for each target, whereas the third row serves as a reference where the inputs of this profile are shown. The linear model has an MSE loss of 1.94 on this operating profile. In general, it can be concluded that the model's performance is satisfactory. Most of the time, the error is under 5°C, which is already acceptable for most commercial applications. The model performs especially well on the NDE bearing temperature with negligible errors. The model does not perform differently between the first half of the time (slow) and the second half (fast) of the operating profile.

### B. Multi-Layer Perception

The second model that has been explored is MLP. As mentioned before, it has the simplest form of a neural network model. However, it is capable of modeling nonlinear relationships rather than linear ones. The MLP layers have been fixed to be two in this work keeping the model structure less complicated. The hyperparameter search proposed a structure with 90 and 20 neurons on each layer. Fig. 8 summarizes the model performance under the same operating profile. The MSE loss for the ML model is 0.31. The performance is much better than that of the linear model, as expected. This model greatly reduces the offset at the end of the first half of the operating profile. It shows an obvious difference between slow and fast dynamics.

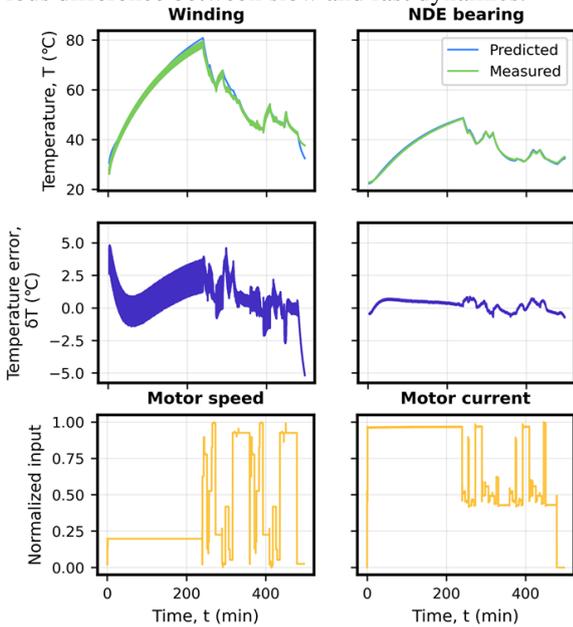

Fig. 7. Linear model performance under the operating profile of Fig. 6.

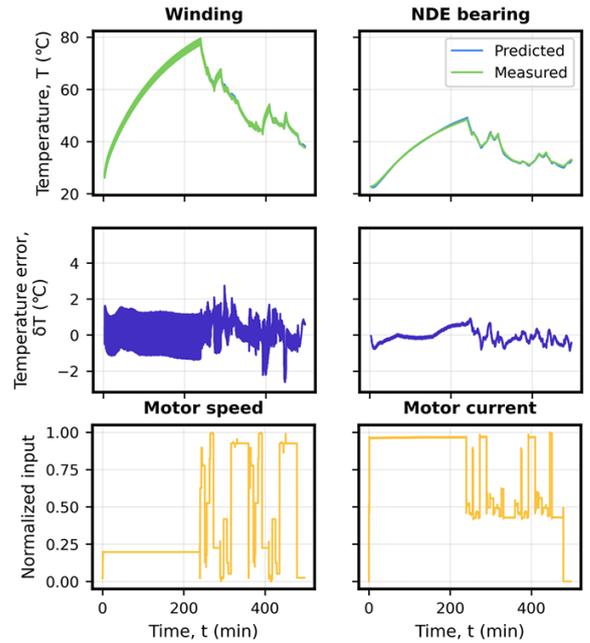

Fig. 8. Multilayer perceptron model performance under the operating profile of Fig. 6.

### C. Convolutional Neural Networks

CNN is the only sequential model that has been tested in this work. Due to its nature, it is expected to learn from the sequential nature of the temperature data faster. However, the model is more complicated and significantly larger than the previous two. In the hyperparameter search, the number of layers has been fixed to three. Relatively small filter sizes and a large dilation rate on the first layer have been found by the search. Fig. 9 gives the results of the CNN model under the same operating profile. The MSE losses are 0.54, while the model prediction of the motor targets was marginally worse than the MLP model. Most of the time, the temperature remained smaller than 3°C.

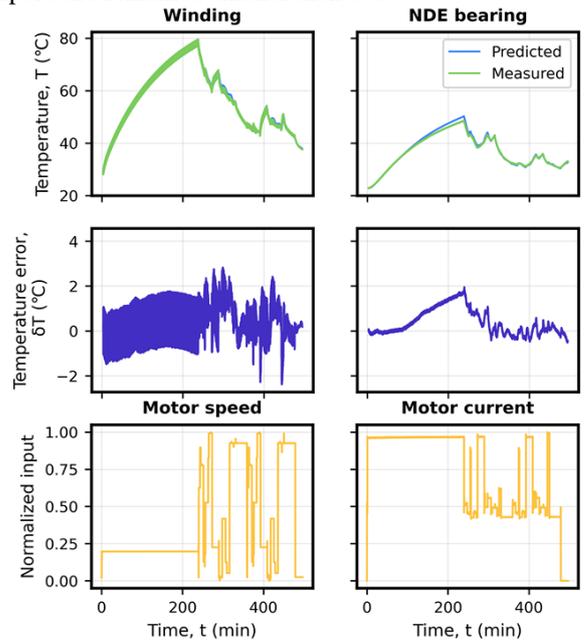

Fig. 9. CNN model performance under the operating profile of Fig. 6.

## VI. Conclusion

In this paper, a machine-learning approach has been applied to estimate internal motor temperatures. Multiple machine learning algorithms were tested on their ability to estimate the temperatures of the stator winding and bearing. The investigated algorithms ranged from an ordinary least square to a deep neural network. For this purpose, experimental lab data has been recorded by letting the system run under predefined operating conditions with slow and fast dynamics. A hyperparameter search was then conducted for each model to find the best configuration. All the algorithms were evaluated by several metrics, and it was found that neuron networks had a satisfactory performance even under fast transient conditions.